\documentclass[review]{elsarticle}

\usepackage{lineno,hyperref}
\usepackage{ifthen}  
\usepackage{multicol}   
\usepackage{amssymb,tabulary,dsfont}                       
\usepackage[ruled,linesnumbered]{algorithm2e}
\usepackage{graphicx,subfigure,booktabs,epsfig}
\usepackage{mdwmath}
\usepackage{mdwtab,amsmath,amsthm,multirow,mathrsfs}
\usepackage{rotating}
\usepackage{slashbox}
\usepackage{epstopdf}
\usepackage{color}

\journal{Digital Signal Processing}

\begin{document}

\begin{frontmatter}

\title{Subsampled terahertz data reconstruction based on spatio-temporal dictionary learning}

\author[vahid]{Vahid Abolghasemi}
\address[vahid]{Faculty of Electrical Engineering, University of Shahrood, Shahrood, 3619995161, Iran}
\cortext[vahid]{Corresponding author}
\ead{vabolghasemi@ieee.org}

\author[hao]{Hao Shen}
\author[hao]{Yaochun Shen}
\address[hao]{Department of Electrical Engineering and Electronics, University of Liverpool, Liverpool, L69 3GL, UK}

\author[lu]{Lu Gan}
\address[lu]{School of Engineering and Design, Brunel University, Uxbridge, UB8 3PH, UK}

\begin{abstract}
In this paper, the problem of terahertz pulsed imaging and reconstruction is addressed. It is assumed that an incomplete (subsampled) three dimensional THz data set has been acquired and the aim is to recover all missing samples. A sparsity-inducing approach is proposed for this purpose. First, a simple interpolation is applied to incomplete noisy data. Then, we propose a spatio-temporal dictionary learning method to obtain an appropriate sparse representation of data based on a joint sparse recovery algorithm. Then, using the sparse coefficients and the learned dictionary, the 3D data is effectively denoised by minimizing a simple cost function.
We consider two types of terahertz data to evaluate the performance of the proposed approach; THz data acquired for a model sample with clear layered structures (e.g., a T-shape plastic sheet buried in a polythene pellet), and pharmaceutical tablet data (with low spatial resolution).
The achieved signal-to-noise-ratio for reconstruction of T-shape data, from only $5\%$ observation was $19$ dB. Moreover, the accuracies of obtained thickness and depth measurements for pharmaceutical tablet data after reconstruction from $10\%$ observation were $98.8\%$, and $99.9\%$, respectively.
These results, along with chemical mapping analysis, presented at the end of this paper, confirm the accuracy of the proposed method.
\end{abstract}


\end{frontmatter}

\section{Introduction}
\label{sec:intro}
Terahertz pulsed imaging (TPI) systems hold great potential in many applications such as medical diagnosis of human tissue, the detection and chemical mapping of illicit drugs and explosives, and pharmaceutical tablet inspection \cite{Ychen_tablet1,ychen_tablet3,Muller2012690}.  In a typical TPI measurement, the THz waveform for each pixel is recorded as a function of optical time delay. This provides a three-dimensional (3D) data set where two axes describe an object's spatial content, and the third one represents the (time) depth information. In TPI, the transient electric field, rather than the radiation intensity, is measured. Thus, by applying the one dimensional Fourier transform along the time axis we can get the magnitude and phase information of THz spectral data for each pixel. Despite these advantages, most existing TPI systems suffer from slow scanning speed and high implementation cost due to pixel-by-pixel raster scan mechanism. One approach to speed up the acquisition process is to reduce the number of samples (measurements) and yet preserve the reconstruction quality. Recently, compressive sampling (CS) \cite{CS_Donoho,CS_Candes1580791} has been widely used for this purpose. This theory essentially works based on random projections of input signals and sparse representation \cite{CS_intro} during reconstruction. It holds great potential for sampling rates reduction, imaging time, power consumption and computational complexity. However, the main challenge in utilizing CS is implementation costs and specifically designing the sampling operator which requires costly and sophisticated hardware modules.
Most recent researches on CS-THz has been focused to address this issue. For instance, in \cite{Hao_spin} the authors propose a spining disk as a plausible sampling operator for high-speed compressive acquisition. In \cite{block_CS2}, a block-based CS method is proposed, and in \cite{Chan2008} a single pixel camera based on Bernoulli random matrix is presented. Recently, an improved version of \cite{block_CS2}, called adaptive CS, is reported \cite{adaptiveCS}. In this method, additional measurement points are adaptively added at the regions prone to degradation with the aim of improving reconstruction quality.

In this paper, we consider a very simple acquisition scenario, i.e. incomplete (subsampled) data, which can be easily realized without too much of hardware burden. Instead, we aim to utilize advanced reconstruction techniques to retrieve the original data samples with least possible quality degradation. In the sequel, more details are explained.

It should be noted that the spatial and temporal (or spectal) features in 3D TPI data have intrinsic geometrical structures. If these structures are modelled properly, it may lead to new TPI imaging systems with faster acquisition speed and more accurate data analysis. This aspect is an open issue and has not been specifically studied in the THz literature. Over the past few years, there have been increased interests in the study of  sparse representation, in which a signal is characterized by a few non-zero coefficients in a certain transform domain.
Up until now, sparse modelling has found applications in many image processing-related tasks such as acquistion, denoising, inpainting and super-resolution \cite{inpaint1,inpaint2,inpaint3,superres1,superres2}. Most of existing works are limited to a deterministic sparsifying transform, such as the Fourier, the wavelet and the curvelet etc. More recently, it has been shown that a signal can be represented with fewer coefficients over a learned dictionary from a number of training samples (see \cite{KSVD} and references therein).

In this paper, we aim to investigate dictionary learning from incomplete and noisy 3D TPI data set. In particular, the missing data are first estimated from random subsets using Bicubic interpolation. Incomplete TPI data set can occur in two different scenarios, in which both require retrieving the missing samples; \emph{i}) result of data transfer from a sender to receiver, and \emph{ii}) subsampled THz data: a simplified case of CS where merely some samples are missing (this obviously requires much simpler sampling operator than original CS).
As a novel tool to address the sample recovery problem, 3D dictionary learning is used here. Spatial-temporal dictionary learning is applied to the 3D data set by exploiting the \emph{joint} sparse model. Finally, denoising is performed based on the learned dictionary through convex optimization. Experimental results show that even with 5\% to 20\% subsampled data, one can still get reliable spatial, structural and spectral information of the object.  To the best of our knowledge, this is the first work demonstrating the advantages of 3D dictionary learning for THz data. These results can be exploited to speed up TPI measurement process substantially by reducing the quantity of data.

 This paper is organized as follows. In the next section we describe the proposed method including mathematical expression of the model, dictionary learning and joint sparse recovery. Section \ref{sec:results} is devoted to demonstrating its applicability to the experimental results. Finally, the conclusion is drawn in section \ref{sec:conclu}.

\section{Proposed method}
\label{sec:proposed}

\subsection{Problem formulation}
\label{sec:problem}

\begin{figure}[!t]
\centering
\includegraphics[width=12cm]{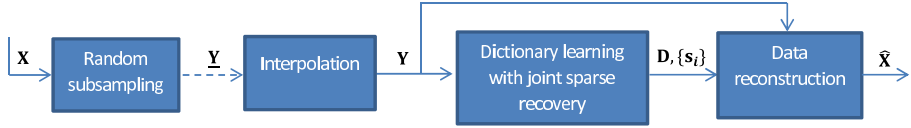}\\
\vspace{-.2cm}
  \caption{Block diagram of the proposed approach.}\label{fig:proposed}
\end{figure}

The first step for recovering the original data from incomplete (subsampled) observations is inpainting. This can be simply achieved using a preliminary and fast approach such as \emph{Bicubic} interpolation \cite{block_CS2}. The main drawback is that the results of such techniques are of low quality especially in noisy scenarios. Other inpainting methods which rely on fixed transforms (e.g. 3D wavelet) \cite{3D-wavelet} also do not lead to acceptable results as we will see later in our experimental results. A recent family of methods which try to build up an adaptive dictionary directly over the incomplete data has shown improved results for 2D natural images when the observation rate is above $25\%$ \cite{elad_sparse_book}. Efficient extension of these techniques for THz data, especially in severely subsampled noisy data ($<25\%$) is our objective. To take advantages of existing techniques for both incomplete and noisy data we propose a new model shown in Fig. \ref{fig:proposed}. If we denote the original clean THz datacube by $\mathbf{X}\in\mathbb{R}^{N_x\times N_y\times B}$ and the incomplete noisy datacube by $\underline{\mathbf{Y}}$ (of the same size), the estimated low-quality noisy datacube can be represented by $\mathbf{Y}$ (of the same size). Our aim is to denoise $\mathbf{Y}$ using a dictionary which is specifically designed to exploit both spatial and temporal correlations existing within THz data.

\subsection{Dictionary learning}
\label{sec:diclearn}

\begin{figure}[!t]
\centering
\includegraphics[width=8cm]{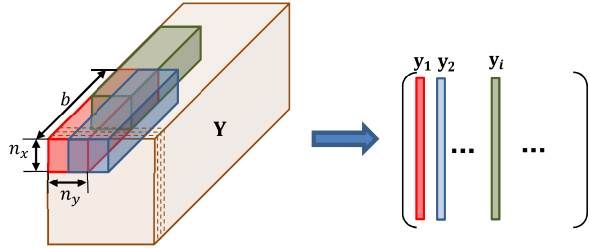}\\
\vspace{-.2cm}
  \caption{Geometric illustration of block extraction procedure from the THz datacube.}\label{fig:voxel}
\end{figure}

Since the THz data is huge, one cannot learn a dictionary directly over the entire datacube. More importantly, 3D nature of THz data should be taken into account when constituting the training samples.
Assume that $\mathbf{y}_i$ is the $i$-th training vector which is obtained through partitioning $\mathbf{Y}$ into $N$ small 3D blocks of size $n_{x}\times n_{y}\times b$, where, $(n_x,n_y)$ and $b$ are the block sizes in spatial and temporal dimensions, respectively. These blocks may have spatial and temporal overlaps and are always converted to the column-vector $\mathbf{y}_i$ of length $r=n_xn_yb$. Fig. \ref{fig:voxel} illustrates the geometrical view of the proposed block extraction. In addition to capturing spatio-temporal structure in the proposed scheme, we have the flexibility to define the contributions of spatial and temporal dimensions by adjusting $(n_x,n_y)$ and $b$, respectively.

The next step is to find the dictionary $\mathbf{D}\in\mathbb{R}^{r\times k}$ and sparse coefficients, denoted by $\{\mathbf{s}_i\}_{i=1}^N$, from the training set $\{\mathbf{y}_i\}_{i=1}^N$. In spite of most traditional dictionary learning methods in which all training vectors are treated independently, we devote a different approach. In order to effectively exploit the 3D spatio-temporal structure of THz data we cumulate every $l$ training vectors into a matrix $\mathbf{\Omega}_j\in\mathbb{R}^{r\times l}$ and solve the following problem:
\begin{equation}\label{eq:MMV}
\mathbf{\Omega}_j=\mathbf{D}\mathbf{S}_j+\mathbf{E}_j,
\end{equation}
\noindent where $j$ denotes the $j$-th subset, $\mathbf{D}\in\mathbb{R}^{r\times k}$ is the dictionary, $\mathbf{S}_j\in\mathbb{R}^{k\times l}$ is the corresponding sparse coefficients matrix, and $\mathbf{E}_j\in\mathbb{R}^{r\times l}$ represents the decomposition error. This is a joint sparse model which simultaneously
deals with the training vectors in $\mathbf{\Omega}_j$.
There are different ways to choose the training subset $\mathbf{\Omega}_j$. For instance, one can cluster the 3D blocks and then group them based on the amount of similarity within the voxels of these blocks. For simplicity, and yet preserving the sparsity structure of neighboring blocks, we use raster-scanning with full spatial and temporal overlap from left-to-right and top-to-bottom as shown in Fig. \ref{fig:voxel}. We have observed encouraging results using this scheme, but the performance of more sophisticated grouping techniques can be further investigated in the future. Here, we group all the blocks into $\lceil N/l\rceil$ subsets of size $l$:
\begin{equation}\label{eq:omega}
\underbrace{\{\mathbf{y}_1,\ldots\,\mathbf{y}_l\}}_{\mathbf{\Omega}_1},
\underbrace{\{\mathbf{y}_{l+1},\ldots\,\mathbf{y}_{2l}\}}_{\mathbf{\Omega}_2},
\ldots\underbrace{\{\mathbf{y}_{N-l+1},\ldots\,\mathbf{y}_N\}}_{\mathbf{\Omega}_{\lceil N/l\rceil}}, \nonumber
\end{equation}
\noindent where $\lceil\cdot\rceil$ is the ceiling operator.
Correspondingly, $\mathbf{S}_j$'s are defined as $k\times l$ matrices as follows:
\begin{equation}
\underbrace{\{\mathbf{s}_1,\ldots\,\mathbf{s}_l\}}_{\mathbf{S}_1},
\underbrace{\{\mathbf{s}_{l+1},\ldots\,\mathbf{s}_{2l}\}}_{\mathbf{S}_2},
\ldots\underbrace{\{\mathbf{s}_{N-l+1},\ldots\,\mathbf{s}_N\}}_{\mathbf{S}_{\lceil N/l\rceil}}, \nonumber
\end{equation}
and the aim is to solve the following minimization problem subject to sparsity of $\mathbf{S}_j$:
\begin{equation}\label{eq:costfunc1}
\forall j \;\min_{\mathbf{D},\mathbf{S}_j}\left\|\mathbf{\Omega}_j-\mathbf{D}\mathbf{S}_j\right\|_F^2
\end{equation}
Most typical dictionary learning approaches work based on \emph{``alternating minimization''} \cite{KSVD,MOD1257971,olshausen.and.field.1996}.
These methods mainly contain two major steps, coefficient update and dictionary update, which are alternately executed until reaching local minima for (\ref{eq:costfunc1}).

\begin{figure}[!t]
\centering
  \includegraphics[width=8cm]{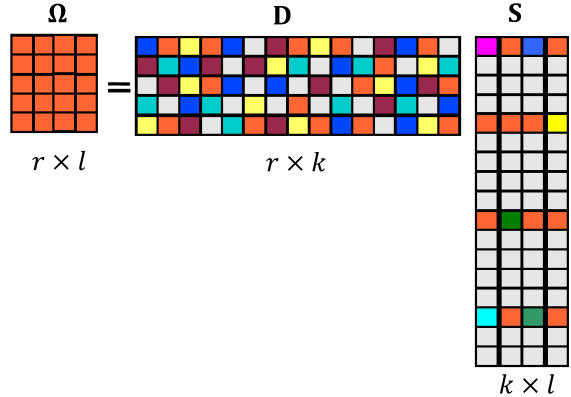}
  \vspace{-.5cm}
  \caption{MMV model}\label{fig:diagram}
\end{figure}

In order to jointly update the sparse coefficients we choose MMV (multiple measurement vector) sparse coding framework. In this framework, as opposed to SMV (single measurement vector), several sparse vectors are simultaneously recovered. The main assumption in MMV is that the sparse vectors admit a common sparsity pattern, i.e., the locations of non-zeros are the same for all vectors. Fig. \ref{fig:diagram} is a diagram illustrating the MMV model.
There are several algorithms in the literature to solve (\ref{eq:costfunc1}) with respect to $\mathbf{S}_j$. Most of these methods are extended from SMV-based approaches \cite{M_FOCUSS,SOMP1,SOMP2}. We have found that SOMP\footnote{Available in SPAM: \url{http://spams-devel.gforge.inria.fr/doc/html/doc_spams002.html}.} (Simultaneous Orthogonal Matching Pursuit) \cite{SOMP1,SOMP2} works well for reconstruction of THz data and hence we utilized it for coefficients update.

Next, for the dictionary update (solving (\ref{eq:costfunc1}) with respect to $\mathbf{D}$), we use a well-established method called K-SVD \cite{KSVD}. K-SVD\footnote{KSVD-Box: \url{http://www.cs.technion.ac.il/~ronrubin/software.html}} is a generalization of K-means clustering and updates the dictionary in a column-by-column scheme by using singular value decomposition (SVD) (full details in \cite{KSVD}).

\subsection{THz data reconstruction}
\label{sec:denoising}

After finding $\mathbf{D}$ and $\{\mathbf{s}_i\}_{i=1}^N$, we need to estimate the THz data. To do this, we define the following minimization problem:
\begin{align}\label{eq:inpaint}
\min_{\mathbf{x}}\lambda\left\|\mathbf{y}-\mathbf{x}\right\|_2^2+
\sum_{i=1}^{N}\left\|\mathbf{R}_i\mathbf{x}-\mathbf{D}\mathbf{s}_i\right\|_2^2+
\beta\sum_{i=1}^N\left\|\mathbf{R}_i\mathbf{x}-\mathbf{m}_{i}\right\|_2^2.
\end{align}
\noindent where $\mathbf{y}=\textrm{vec}(\mathbf{Y})$ and $\mathbf{x}=\textrm{vec}(\mathbf{X})$, both with length $p=N_xN_yB$, are vectorized representations of datacubes $\mathbf{Y}$ (noisy) and $\mathbf{X}$ (clean), respectively.\footnote{Note that $\mathbf{x}$ and $\mathbf{y}$ should not be confused with 3D blocks where always have subscript indices thorough the paper.} Also, $\mathbf{y}_i=\mathbf{R}_i\mathbf{y}$ and $\mathbf{x}_i=\mathbf{R}_i\mathbf{x}$, where $\mathbf{R}_i$ is a huge binary matrix of size $r\times p$. $\mathbf{R}_i$ has only one non-zero (i.e. 1) in each row. As a block extraction operator, $\mathbf{R}_i$ extracts the voxels belonging to $i$-th block.

In (\ref{eq:inpaint}), the leftmost term is the error between noisy and clean data, the middle term is related to the sparse decompostion error, and the rightmost term is the smoothness constraint with regularization parameter $\beta$. Moreover, $\mathbf{m}_{i}=[m_{i},m_{i},\ldots]^T$ is the $r\times1$ mean vector of the $i$-th block (i.e. average of all elements in $\mathbf{R}_i\mathbf{x}$). Problem (\ref{eq:inpaint}) is convex in $\mathbf{x}$ and can be solved by zeroing its gradient, with respect to $\mathbf{x}$, which finally leads to:
\begin{equation}\label{eq:X_update_2}
\widehat{\mathbf{x}}=\left(\lambda\mathbf{I}+(1+\beta)\sum_{i=1}^N{\mathbf{R}_i^T\mathbf{R}_i}\right)^{-1}
\left(\lambda\mathbf{y}+\sum_{i=1}^N{\mathbf{R}_i^T\left(\mathbf{D}\mathbf{s}_{i}+\beta\mathbf{m}_{i}\right)}\right).
\end{equation}
\noindent Although the above expression seems computationally expensive at the first glance, it does not need to be directly calculated in practice. Instead, since the inverting term in (\ref{eq:X_update_2}) is diagonal, we obtain the estimated datacube in a voxel-wise fashion (similar to the strategy used in \cite{denoise_Elad}). We can show the voxel-wise calculation of one voxel $\hat{x}$ (taking the corresponding voxels in $\mathbf{y}$ and $\mathbf{m}$ into account) using the following sets of equations:
\begin{align}\label{eq:inpaint_voxel}
\min_{x}\left\{\mathcal{C}(x)=\lambda(x-y)^2+(x-ds)^2+\beta(x-m)^2\right\}\\ \nonumber
\rightarrow \frac{\partial\mathcal{C}}{\partial x}=2\lambda(x-y)+2(x-ds)+2\beta(x-m)=0\\ \nonumber
\rightarrow \hat{x}=\frac{1}{1+\beta+\lambda}\left(\lambda y+ds+\beta m\right)
\end{align}
\noindent where $\mathcal{C}$ is the cost function to be minimized, and $ds$ is the corresponding voxel in $\mathbf{Ds}$. It is seen that the last equation above well matches with (\ref{eq:X_update_2}) which is the non-practical form of reconstructing datacube.

The pseudo-code of the proposed method is given in Algorithm \ref{alg:adaptive_mmca}.

\begin{algorithm}[!t]
\small
\caption{Pseudo-code of the proposed method.}\label{alg:adaptive_mmca}
\KwIn{Incomplete noisy datacube $\underline{\mathbf{Y}}$.}
\KwOut{Recovered datacube $\widehat{\mathbf{X}}$.}
\Begin{
\emph{Interpolation}:\\
Apply \emph{Bicubic} interpolation on $\underline{\mathbf{Y}}$ and yield $\mathbf{Y}$\;
\emph{Dictionary learning}:\\
Extract the spatio-temporal blocks from $\mathbf{Y}$ and obtain $\{\mathbf{y}_i\}_{i=1}^N$\;
Group the blocks and constitute $\{\mathbf{\Omega}_j\}_{j=1}^{\lceil N/l\rceil}$\;
Initialize the dictionary $\mathbf{D}$ with the DCT (discrete cosine transform) basis\;
\Repeat{stopping criterion is met}{
Apply SOMP to $\mathbf{\Omega}_j$ and yield $\mathbf{S}_j$ for all $j=1,\ldots\lceil N/l\rceil$\;
Update $\mathbf{D}$ using K-SVD dictionary update \cite{KSVD}\;
}
\emph{THz data reconstruction}:\\
Calculate (\ref{eq:X_update_2}) to yield $\widehat{\mathbf{x}}$ and then reshape it to $\widehat{\mathbf{X}}$\;
}
\end{algorithm}

\section{Experimental results}
\label{sec:results}

We present the results of applying the proposed method to two different types of THz data. The first dataset, which we
call it \emph{``T-shape''}, has size $200\times200\times512$, and acquired across an area
of 20mm$\times$22mm using a TPIscan-1000 system (TeraView Ltd, Cambridge, U.K.), covering a spectral range from 0.1 to 3.5 THz.
The sample used is a polythene pellet of a diameter of 25 mm. Inside the pellet there is a T-shaped plastic sheet which locates approximately 0.2 mm below the sample surface. For this dataset, the \emph{structural} information such as thickness and depth are of interest.
The second data is a low spatial resolution data called \emph{``Tablet data''} acquired from pharmaceutical tablets. Two of such sets (called LA and TP) have size $42\times42\times512$, and other seven sets have size $49\times49\times609$. Extraction of \emph{spectral} information such as chemical mapping is crucial for this dataset.

\begin{figure}[!t]
\centering
\subfigure[]{\includegraphics[width=3.4cm]{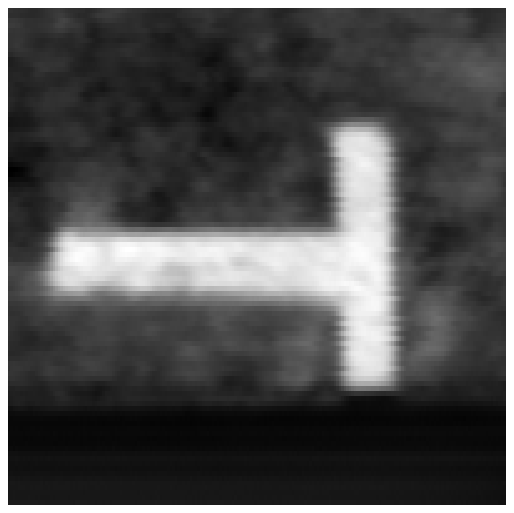}}
\subfigure[]{\includegraphics[width=3.4cm]{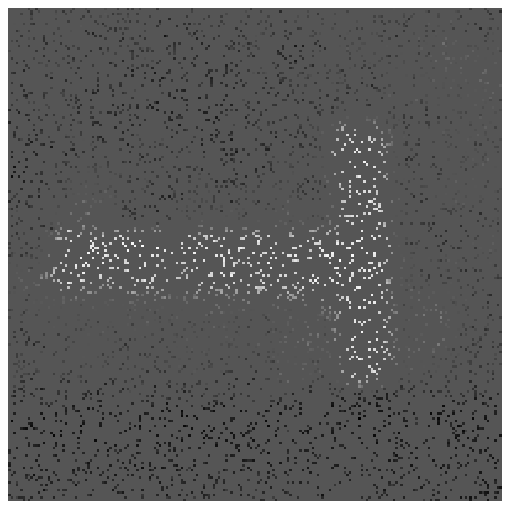}}\\
\subfigure[]{\includegraphics[width=3.4cm]{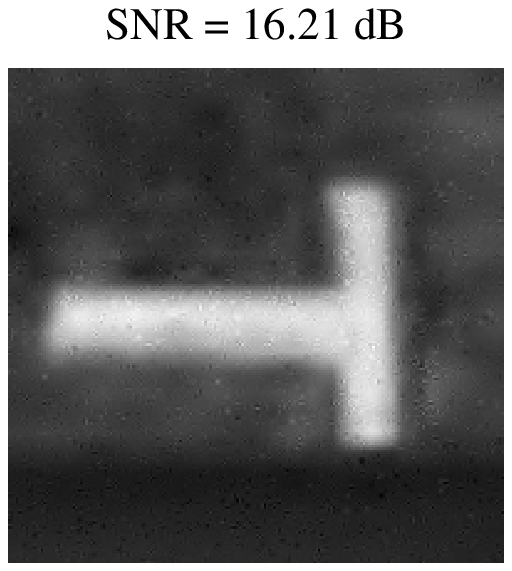}}
\subfigure[]{\includegraphics[width=3.4cm]{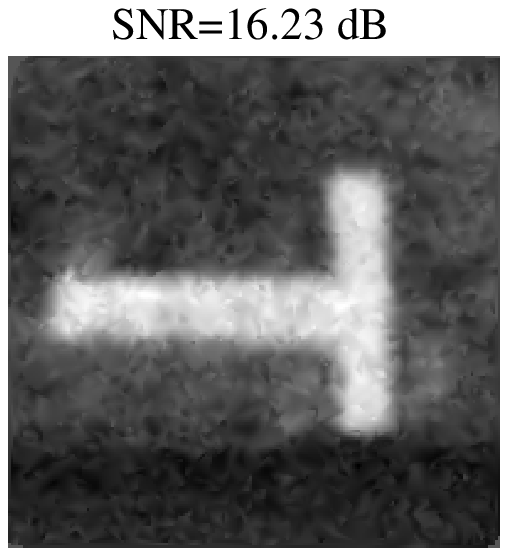}}
\subfigure[]{\includegraphics[width=3.4cm]{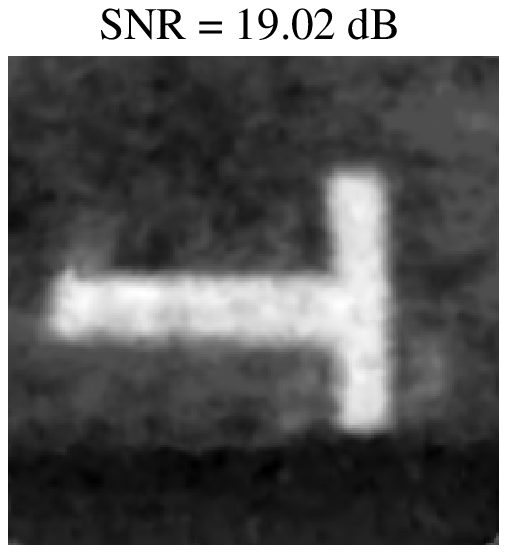}}
\vspace{-.2cm}
  \caption{T-shape data at one temporal band: (a) original image, (b) $10\%$ incomplete noisy observations. Reconstruction results using (c) soft-thresholding in 3D wavelet domain, (d) \emph{Bicubic} interpolation, and (e) proposed method.}\label{fig:incomplete_152}
\end{figure}

\subsection{T-Shape data}

\begin{figure}[!t]
\centering
\includegraphics[width=8cm]{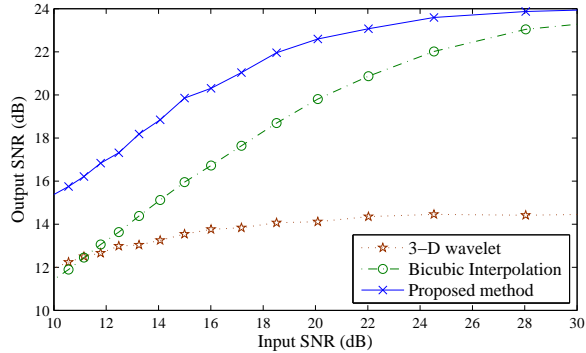}
\vspace{-.2cm}
  \caption{T-shape data: SNR of reconstructed data at $10\%$ observation rate versus input SNR.}\label{fig:SNR_Tshpae}
\end{figure}

Following the proposed model, we first applied \emph{Bicubic} interpolation to the noisy incomplete data, yielding $\mathbf{Y}$. We considered input additive Gaussian noise, leading to input SNR in the range $15\sim20$ dB, which is a reasonable range in practice.
We then applied the proposed dictionary learning method to find the dictionary and corresponding sparse coefficients to reconstruct the entire datacube. Due to importance of structural information in T-shape data, we gave more contribution to the spatial domain than temporal domain for the dictionary learning. Hence, the experiments were conducted by setting the spatial size $n_x=n_y=8$, and two different temporal sizes: $b=1$ (spatial dictionary), and $b=4$ (spatio-temporal dictionary). In any case we used full spatial and temporal overlaps. According to these settings, the obtained dictionaries were of size $64\times256$ and $256\times256$. A sample reconstructed image with $b=4$, $l=10$, $\lambda=0.5$, $\beta=0.1$, and input SNR of $17$ dB is given in Fig. \ref{fig:incomplete_152}. It is seen from this figure that the proposed method has recovered and denoised the original image from only $10\%$ of noisy samples with highest SNR among other methods. The proposed method was able to enhance the SNR of \emph{Bicubic} interpolation results by up to $2.5$ dB. Other methods, i.e. soft-thresholding after applying 3D \emph{symlet4} wavelet transform \cite{3D-wavelet}, shows weaker performances compared to the proposed method. We also show in Fig. \ref{fig:SNR_Tshpae} the reconstructed SNR at the fixed observation rate of $10\%$ versus different input SNRs. As seen from this figure, the proposed approach outperforms other methods.

Next, performance of the proposed method against different values of $b$ and $l$ is investigated. For this purpose, SNRs of the recovered datacube for different observation rates are given in Table \ref{table1}. From this table, the advantage of spatio-temporal dictionary can be observed by comparing the SNRs at $b=4$, with those at $b=1$. Also, the given SNRs in this table indicates that the proposed method is more effective than 3D wavelet transform. We empirically observed that $l\approx10$ gives the best performance for T-shape dataset.

\begin{table}[!t]
\footnotesize
   \caption{T-shape data: recovery accuracy (SNR) in dB for different methods. The results are shown for several observation rates, and different selections of $b$ and $l$.}
   \label{table1}
   \centering
\begin{tabular}{|c|c|c|c|l|}
  \hline
  \multirow{2}{*}{observation rate} & \multicolumn{3}{c|}{SNR (dB)} & \multirow{2}{*}{$b$, $l$} \\ \cline{2-4}
  & 3D wavelet & \emph{Bicubic} interp. & proposed method & \\ \hline
   \multicolumn{1}{|c|}{\multirow{4}{*}{$5\%$}}
    & \multirow{4}{*}{10.53} & \multirow{4}{*}{16.89} & 17.45 & 1, 1 \\ \cline{4-5}
   \multicolumn{1}{|c|}{} &   & &17.69 &1, 10\\ \cline{4-5}
   \multicolumn{1}{|c|}{} &  & & 18.78 &4, 1   \\ \cline{4-5}
   \multicolumn{1}{|c|}{} &  & & 18.96 &4, 10 \\ \hline \hline
  \multicolumn{1}{|c|}{\multirow{4}{*}{$10\%$}}
   & \multirow{4}{*}{14.71} & \multirow{4}{*}{17.68} &19.41 & 1, 1 \\ \cline{4-5}
   \multicolumn{1}{|c|}{} & & &19.46&1, 10\\ \cline{4-5}
   \multicolumn{1}{|c|}{} &  & &20.17 &4, 1 \\ \cline{4-5}
   \multicolumn{1}{|c|}{} &  & & 20.94&4, 10 \\ \hline \hline
   \multicolumn{1}{|c|}{\multirow{4}{*}{$15\%$}}
    & \multirow{4}{*}{16.72} &  \multirow{4}{*}{17.87} & 20.05 & 1, 1 \\ \cline{4-5}
   \multicolumn{1}{|c|}{}  &  && 20.33&1, 10  \\ \cline{4-5}
   \multicolumn{1}{|c|}{}  &  & & 20.86&4, 1  \\ \cline{4-5}
   \multicolumn{1}{|c|}{}  &  & & 22.25&4, 10 \\ \hline \hline
   \multicolumn{1}{|c|}{\multirow{4}{*}{$20\%$}}
    & \multirow{4}{*}{17.69} & \multirow{4}{*}{17.96} & 20.88& 1, 1 \\ \cline{4-5}
   \multicolumn{1}{|c|}{}  &  & &21.03 &1, 10  \\ \cline{4-5}
   \multicolumn{1}{|c|}{}  &  && 22.53&4, 1  \\ \cline{4-5}
   \multicolumn{1}{|c|}{}  &  && 23.15&4, 10 \\ \hline
\end{tabular}
\vspace{-.2cm}
\end{table}

Table \ref{table_time} gives an insight about the computational complexity of the proposed method. In this table, the elapsed time of the dictionary learning stage for different selections of $b$ and $l$ is shown. It is clearly seen that large values of $l$ can significantly reduce the computation time. This behavior supports the advantages of using joint sparse recovery which has already shown to improve the quality as well (in Table \ref{table1}). It can be further seen from Table \ref{table_time} that increasing $b$ also decrease the computation time, though not as dramatically as that for $l$.

\begin{table}[!t]
\footnotesize
   \caption{T-shape data: The computation time of dictionary learning step in the proposed method for different selections of $b$ and $l$. These times are in second and calculated per frame.}
   \label{table_time}
   \centering
\begin{tabular}{|l|l|l|}
  \hline
  \backslashbox{$b$, $l$\kern-3em}{\kern-3em observation\\ rate}
   & $10\%$ & $20\%$\\ \hline
   1,1 & 110.3 & 122.5 \\ \cline{1-3}
   1, 10  &5.0  & 7.6 \\ \cline{1-3}
    4, 1  &93.3  & 97.6      \\ \cline{1-3}
    4, 10  &4.8 & 5.9 \\ \hline
\end{tabular}
\vspace{-.2cm}
\end{table}

\begin{figure}
\centering
\subfigure[]{\includegraphics[width=4.9cm]{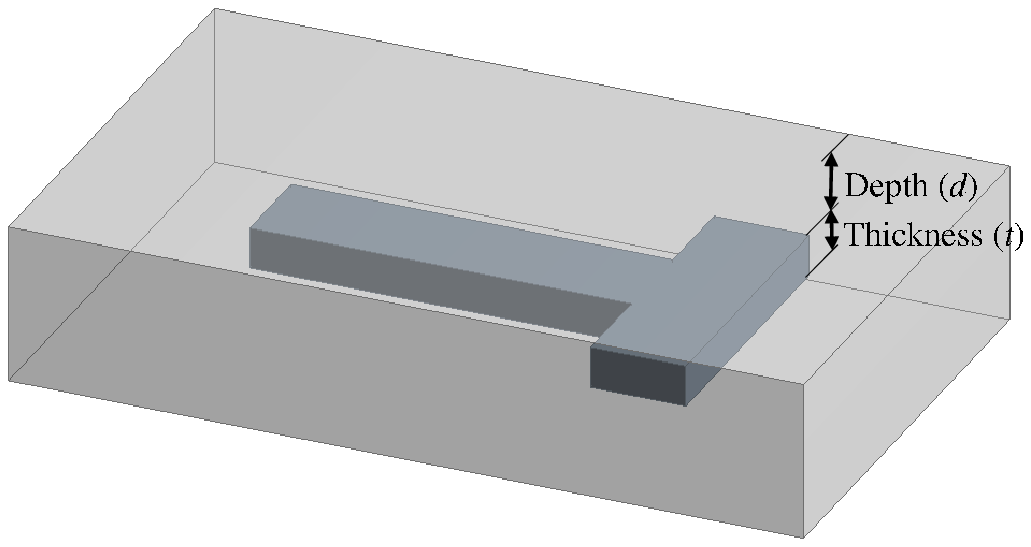}}
\subfigure[]{\includegraphics[width=7cm]{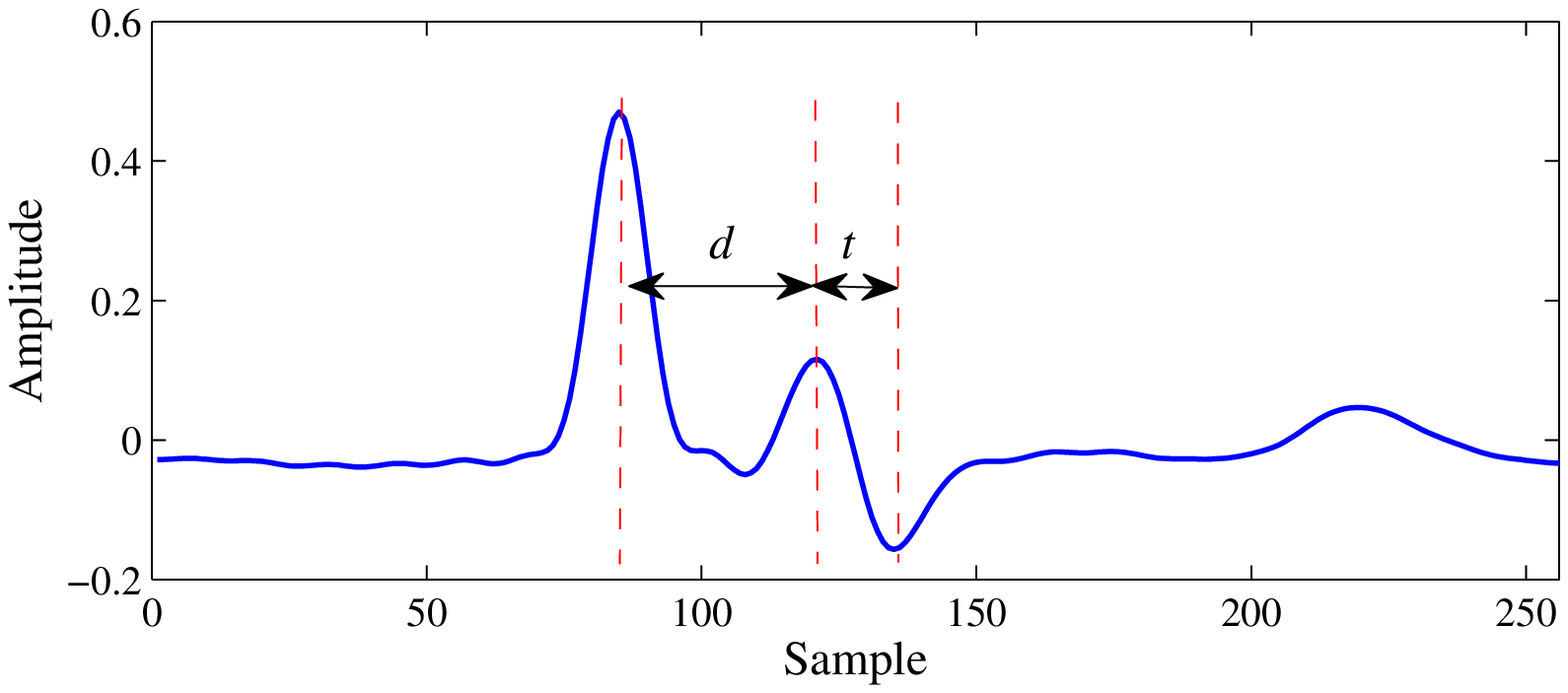}}
\vspace{-.1cm}
  \caption{T-shape data: (a) geometrical representation of T-shape object, and (b) sample time-domain waveform at one spatial location.}\label{fig:wave_tshape}
\end{figure}

As a useful measure, we performed the \emph{thickness} and \emph{depth} evaluations before and after reconstruction of T-shape object. These results give structural information about the object of interest.
Fig. \ref{fig:wave_tshape} (a) geometrically illustrates these two parameters. Thickness ($t$) and depth ($d$) calculations are practically useful to identify inhomogeneities or defects in the object of interest \cite{Ychen_tablet1}. In order to find these parameters we refer to a sample temporal waveform at one spatial location shown in Fig. \ref{fig:wave_tshape} (b).
The obtained values for these two parameters are subject to appropriate scalings to be converted to millimeter (more details can be found in \cite{Ychen_tablet1}). The calculated means and standard deviations of these parameters for different observation rates are given in Table \ref{table3}. The results in this table are shown for both original and reconstructed data using the proposed method. It is found from Table \ref{table3} that the accuracy of reconstruction using the proposed method is very high, as the depth and thickness are approximately equivalent to the original data.

\begin{table}[!t]
\footnotesize
   \caption{T-shape data: measured ``mean$\pm$standard deviation'' of thickness and depth for different observation rates.}
   \label{table3}
   \centering
\begin{tabular}{c|c|c|c|}
  \cline{2-4}
  & observation rate & reconstructed & original \\ \hline
  \multicolumn{1}{|c|}{\multirow{4}{*}{Thickness (mm)}}
   & $5\%$& $0.0766\pm0.0063$ & \multirow{4}{*}{$0.0750\pm0.0029$} \\ \cline{2-3}
   \multicolumn{1}{|c|}{} &$10\%$& $0.0741\pm0.0063$ &  \\ \cline{2-3}
   \multicolumn{1}{|c|}{} &$15\%$& $0.0756\pm0.0056$ &  \\ \cline{2-3}
   \multicolumn{1}{|c|}{} &$20\%$& $0.0748\pm0.0054$ &  \\ \hline \hline
   \multicolumn{1}{|c|}{\multirow{4}{*}{Depth (mm)}}
   & $5\%$ & $0.1889\pm0.0079$ & \multirow{4}{*}{$0.1902\pm0.0071$} \\ \cline{2-3}
   \multicolumn{1}{|c|}{} &$10\%$ & $0.1909\pm0.0087$ &   \\ \cline{2-3}
   \multicolumn{1}{|c|}{} &$15\%$ & $0.1911\pm0.0079$ &   \\ \cline{2-3}
   \multicolumn{1}{|c|}{} &$20\%$& $0.1899\pm0.0082$ &  \\ \hline
\end{tabular}
\vspace{-.2cm}
\end{table}

\subsection{Tablet data}

The dictionary learning settings for Tablet data is different from those of T-Shape data. Due to very low spatial resolution of Tablet data, it is natural to give a much higher contribution to the temporal dimension during the dictionary learning. This is inline with the fact that the temporal/spectral information is more valuable than spatial information in pharmaceutical studies \cite{Ychen_tablet1}\cite{Ychen_table2_springer}. Therefore, we selected $n_x=n_y=2$ and $b=128$ with full spatial and temporal overlaps within the blocks.

\begin{figure}[!t]
\centering
\includegraphics[width=11cm]{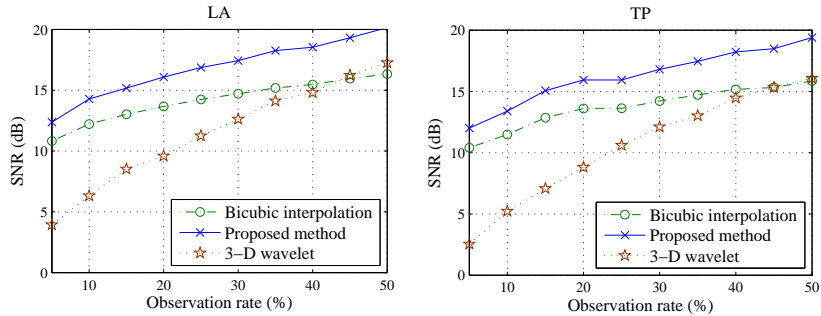}\\
\vspace{-.2cm}
  \caption{Tablet data: the reconstructed SNR versus observation rate for both LA and TP datasets at $20$ dB input SNR.}\label{fig:tablet_SNR}
\end{figure}

As a quantitative measure, we show in Fig. \ref{fig:tablet_SNR} the variations of reconstructed SNR of both LA and TP datasets versus different observation rates, where $l=10$, $k=64$, $\lambda=0.5$ and $\beta=0.2$ were used for the proposed method. These values are chosen empirically and are manually tuned to achieve the best performance among other selections. We also observed that the achieved results are not sensitive to variations of $\lambda$ and $\beta$ around the selected values. The dictionary columns size, i.e. $k$, is set to $64$ due to lower spatial resolution in Tablet data compared to T-shape data. The input SNR for this experiment was set to $20$ dB. Similarly, we show the reconstructed SNR versus input SNR at $20\%$ observation rate. It is clearly seen that the proposed method is able to significantly improve the reconstructed SNR in both figures.

\begin{figure}[!t]
\centering
\includegraphics[width=11cm]{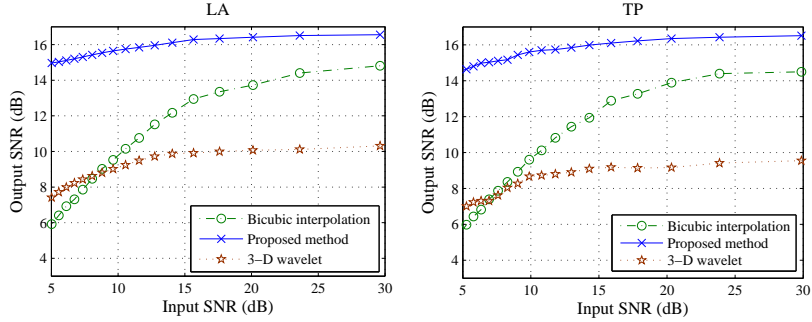}\\
\vspace{-.2cm}
  \caption{Tablet data: the reconstructed SNR versus input SNR for both LA and TP datasets at $20\%$ observation rate.}\label{fig:tablet_SNRinOut}
\end{figure}

In order to observe the robustness of the proposed approach, more sets of tablet data were included in our experiments and the average performance was measured. Seven incomplete Tablet datasets at different observation rates were reconstructed by using well-known methods. Three input SNR levels were used in this experiment. The results are given in Table \ref{table4}. It is seen that the achieved results using the proposed method is better than those obtained using conventional techniques. Moreover, the average output SNRs are compatible with what obtained in the previous experiment (Figures \ref{fig:tablet_SNR} and \ref{fig:tablet_SNRinOut}).

\begin{table}[!b]
\footnotesize
   \caption{Tablet data: output SNRs (dB) at different observation rates and input SNRs (dB), averaged over seven datasets.}
   \label{table4}
   \centering
\begin{tabular}{|c|c|c|c|c|}
  \hline
    \multirow{2}{*}{ Input SNR (dB)} &   \multirow{2}{*}{observation rate ($\%$)}& \multicolumn{3}{c|}{Output SNR (dB)} \\ \cline{3-5}
 & & 3D wavelet & \emph{Bicubic} interp. & proposed method \\ \hline
  \multicolumn{1}{|c|}{\multirow{3}{*}{10}}
   & $10$& 7.91 & 9.49 & 14.71 \\
   \multicolumn{1}{|c|}{} &$20$& 9.31 & 11.14 & 16.26 \\
   \multicolumn{1}{|c|}{} &$30$& 12.16 & 13.75 & 18.12 \\ \hline
   \multicolumn{1}{|c|}{\multirow{3}{*}{20}}
   & $10$ & 8.34 &  12.40 & 16.13 \\
   \multicolumn{1}{|c|}{} &$20$ & 10.67 & 14.83 & 17.22   \\
   \multicolumn{1}{|c|}{} &$30$& 12.24 & 15.91 & 19.49  \\ \hline
   \multicolumn{1}{|c|}{\multirow{3}{*}{30}}
   & $10$ & 9.81 & 14.92 & 18.23 \\
   \multicolumn{1}{|c|}{} &$20$ & 12.64 & 16.78 & 20.12   \\
   \multicolumn{1}{|c|}{} &$30$& 14.31 & 18.63 & 22.70  \\ \hline
\end{tabular}
\end{table}

\begin{figure}[!t]
\centering
\subfigure[]{\includegraphics[width=11cm]{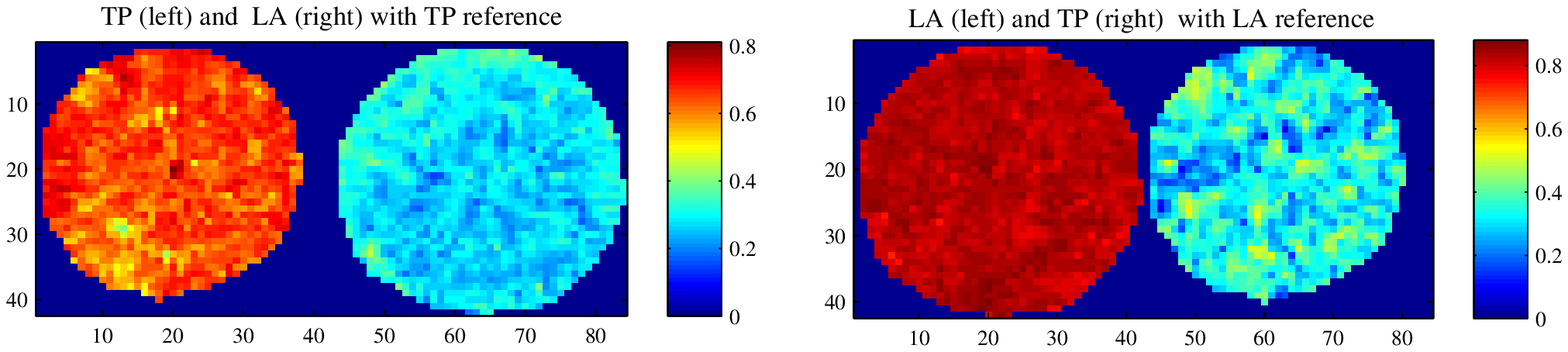}}
\subfigure[]{\includegraphics[width=11cm]{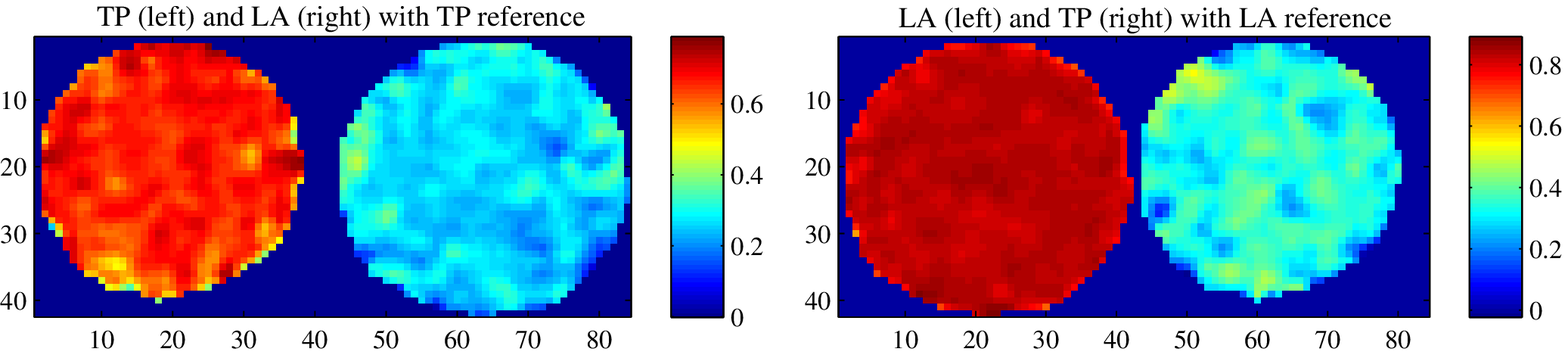}}
\vspace{-.2cm}
  \caption{Tablet data: the chemical map for (a) original and (b) reconstructed data from $20\%$ observation rate.}\label{fig:tablet}
\end{figure}

As mentioned before, chemical mapping is a useful tool to analyze and observe the uniformity of the tablet \cite{Ychen_tablet1}. Chemical mapping can be obtained using the terahertz spectral information. This evaluation should be performed in the frequency-domain, thus, we first take Fourier transform of the time-domain waveforms for both original and reconstructed data. Then, a reference (spectral) waveform denoted by $\mathbf{u}$ is selected from original LA dataset at one pixel location (normally from the middle of the tablet). After that, a spectral matching should be performed to generate the chemical map. We used the cosine correlation mapping (CCM) \cite{CCM}, as used in \cite{Ychen_tablet1}, for this purpose.
If we define a spectral frequency-domain vector by $\mathbf{v}$, then the CCM can be calculated via:
\begin{equation}\label{eq:CCM}
\cos(\theta)=\sum_{p=1}^P\frac{u_p\cdot v_p}{\left\|\mathbf{u}\right\|_2\cdot\left\|\mathbf{v}\right\|_2},
\end{equation}
\noindent where $P$ is the number of spectral bands. Parameter $\theta$ is the angle between $\mathbf{v}$ and $\mathbf{u}$, and therefore, $\cos(\theta)$ is between 0 and 1. This evaluation should be performed for all pixels of the corresponding dataset. Smaller angle (larger $\cos(\theta)$) means better match between reference spectra and the waveform under evaluation. The results of chemical mapping for our tablet data is shown in Fig. \ref{fig:tablet}. We calculated CCM for four cases; TP with LA reference, LA with LA reference, TP with TP reference, and TP with LA reference. It is seen from Fig. \ref{fig:tablet} that both the reconstructed data have a close chemical map similar to the original ones, and that they have a smooth and uniform shape.

\section{Discussion and Conclusion}
\label{sec:conclu}

In this paper, we addressed the THz data recovery from an incomplete noisy set of observations. The main advantage of starting with incomplete (subsampled) THz data is fast and inexpensive acquisition process. Inspired by recent developments in dictionary learning and sparse recovery frameworks, we proposed a spatio-temporal dictionary learning technique to find an efficient sparse domain for the THz data. The proposed method aimed at exploiting the existing temporal and spatial correlations. It is worthy here to know the difference between the time-domain THz system and visible-light camera by noting their image formation details. In fact, even for a static object, one can still get 3D THz data, while for a visible-light camera, the 3D data leads to a video sequence with moving scene. To be more accurate, in terms of spatial correlation, THz signal exhibits some similar correlations as that obtained by a visible camera. However, in terms of temporal-correlation, it is significantly different from that of video sequences at visible spectrums. Note that the temporal THz data will either give about structural information like thickness and depth or provide some spectral information, e.g., the chemical component of an object, as presented in the previous section.
Extensive sets of experiments were conducted to support the effectiveness and accuracy of the proposed method. The thickness and depth calculations, and chemical mapping analysis for two different types of THz data, along with the achieved SNRs, confirmed the advantages of the proposed approach. In future, we are going to extend the proposed method for the complex scenario where a joint spatio-temporal and frequency domain dictionary learning can be achieved.

\section{Acknowledgement}
This work was supported by the Engineering and Physical Sciences Research Council (EPSRC), UK, under project EP/I038853/1. We would also like to thank Dr. Yue Dong at University of Liverpool for providing new sets of THz data for further analysis and evaluation.

\section*{References}

\bibliography{refs}

\end{document}